\documentclass{article} 
\usepackage{iclr2024_conference,times}

\usepackage{hyperref}
\usepackage{url}
\usepackage{booktabs}
\usepackage{algorithm}
\usepackage[noend]{algpseudocode}
\usepackage{graphicx}
\usepackage{fancyvrb}
\usepackage{wrapfig}
\usepackage{caption}
\usepackage{subcaption}
\usepackage{enumitem}

\usepackage{booktabs}
\usepackage{tabularx}

\usepackage{times}
\usepackage{latexsym}

\usepackage[T1]{fontenc}
\usepackage[utf8]{inputenc}

\usepackage{microtype}

\usepackage{inconsolata}

\usepackage[english]{babel}
\usepackage[utf8]{inputenc}
\usepackage{algorithm}
\usepackage[noend]{algpseudocode}
\usepackage{graphicx}

\title{Bit Cipher — A Simple yet Powerful Word \\ Representation System that Integrates \\Efficiently with Language-Models}

\author{Haoran Zhao \& Jake Ryland Williams 
\\
Department of Information Science, 
Drexel University\\
Philadelphia, PA 19104, USA \\
\texttt{\{hz454,jw3477\}@drexel.edu} \\
}

\iclrfinalcopy 
\begin{document}

\maketitle

\begin{abstract}
While Large Language Models (LLMs) become ever more dominant, classic pre-trained word embeddings sustain their relevance through computational efficiency and nuanced linguistic interpretation. Drawing from recent studies demonstrating that the convergence of GloVe and word2vec optimizations \textit{all} tend towards log-co-occurrence matrix variants, we construct a novel word representation system called  \textbf{\textit{Bit-cipher}} that eliminates the need of backpropagation while leveraging contextual information and hyper-efficient dimensionality reduction techniques based on unigram frequency, providing strong interpretability, alongside efficiency. 
We use the bit-cipher algorithm to train word vectors via a two-step process that critically relies on a hyperparameter---\textit{bits}---that controls the vector dimension. While the first step trains the bit-cipher, the second utilizes it under two different aggregation modes---\textit{summation} or \textit{concatenation}---to produce contextually rich representations from word co-occurrences. We extend our investigation into bit-cipher's efficacy, performing probing experiments on part-of-speech (POS) tagging and named entity recognition (NER) to assess its competitiveness with classic embeddings like word2vec and GloVe. Additionally, we explore its applicability in LM training and fine-tuning. By replacing embedding layers with cipher embeddings, our experiments illustrate the notable efficiency of cipher in accelerating the training process and attaining better optima compared to conventional training paradigms. In fine-tuning experiments, training cipher embeddings on target datasets and replacing the embedding layer of the LMs to be fine-tuned negates the need for extensive model adjustments, offering a highly efficient transfer learning alternative. Experiments on the integration of bit-cipher embedding layers with Roberta, T5, and OPT, prior to or as a substitute for fine-tuning, showcase a promising enhancement to transfer learning, allowing rapid model convergence while preserving competitive performance.

\end{abstract}

\section{Introduction}
\label{introduction}


Word embedding algorithms serve as a crucial tools for understanding the semantics of categorical features in natural language processing (NLP) and deep learning (DL). Moreover, they  continue to form an integral component of modern large language modeling (LLM) systems, since the initial step that LLMs, too, must approach is the efficient representation of tokens by static embeddings.  
Prior to the advent of the transformer architecture, it was research on pre-trained word embedding techniques that enabled DL for NLP. Pioneered by \citet{mikolov2013efficient}, word2vec ushered in NLP's era of representation learning, using the continuous bag-of-words and skip-gram models to demonstrate that it was possible to learn meaningful, low-dimensional representations with limited resources by predicting co-occurring tokens. 

To accelerate learning via summary statistics (co-frequency), GloVe was ultimately introduced to harness the global statistics of co-occurrences~\citep{pennington-etal-2014-glove}, and moreover,
without the use of contrastive learning. From there, it was ultimately a pivot to the modeling of sub-word information in a word2vec-like variant called FastText~\citep{bojanowski2017tricks, bojanowski2017enriching} that guided further pre-transformer advances, leaving us to ask: 
\begin{quote}
Could further improvements instead be made to the objective and optimization of embedding architectures, as opposed to their granularity of application?    
\end{quote} 
Questions like this seem obscure with the arrival  of the transformer, since the research paradigm has shifted from pre-trained word embeddings to more nuanced `contextual' representations, defined by the hidden states of transformers. This shift saw the emergence of powerful models such as BERT, ELMo, and GPT~\citep{devlin2019bert,peters-etal-2018-deep,radford2018improving, radford2019language}, all of which relied on training LLMs to generate even higher-performance representations of words that demonstrate greater nuance at prediction of downstream tasks. However, since transformer embedding layers generally only leverage sub-word information (and positional encoding) over GloVe and word2vec, we see the presented main research question as not only valid, but by extension, capable of improving LLM architectures, since all require some form of embedding.

Notwithstanding the success of LLMs, one should still ask: Does the study of traditional word embedding methods retain any value? In this work, we argue in favor based on the following points: (1) the computational costs of training LLMs are substantial (\citealp{rae2022scaling}, \citealp{thoppilan2022lamda}), and obtaining contextual representations are essentially a by-product rather than the main objective of training LLMs. (2) Due to the cost-intensive nature of training LLMs, there is an inherent non-ideal trade-off between optimal performance and cost-effectiveness in them. 
(3) Initial pre-trained word embedding layers can greatly speed up and/or reduce the costs of training larger models that depend on embedding layers \citep{panahi2020word2ket}.

In this work, we address points (1)--(3) by introducing the \textit{bit-cipher}, which is a technique capable of representing words in a highly efficient manner into user-defined dimensionalities of word vectors. Drawing inspiration from one-hot encoding, the bit-cipher follows a straightforward and explicit process for vector assignment. Moreover, we extend capability by aligning with recent studies showing that the various forms of GloVe and word2vec converge towards variants of $\log$-co-occurrence matrices. While we underscore the efficiency and competitiveness of bit-cipher against other pre-trained word embedding methods, we advise against using it in isolation or comparing it directly with contextual word embeddings. We perceive it more as a means to be used as a component of larger LM architectures, rather than as a standalone utility. In particular, we integrate contextual information via two different methods based on the summation (\textbf{Sum}) and concatenation (\textbf{Cat}) of co-occurrent information.
Our investigations find that concatenation-based models using a large window size perform competitively when compared to GloVe and word2vec on Part-of-Speech (POS) tagging and Named Entity Recognition (NER) tasks, often out-performing both. Furthermore, experiments on integrating cipher embeddings into LM training and fine-tuning are conducted to show two of the main potential use scenarios of bit-cipher stating its efficiency and competitive nature with traditional methods.



\section{Related Work}
\label{related}


\textbf{Generations and types of pre-trained word embeddings: } Representation learning in NLP 
has gone through many large transitions,
starting from static word vectors \citep{mikolov2013efficient,mikolov2013distributed,pennington2014glove},
and into contextual word representations \citep{howard2018universal,peters2018deep},
and now the predominant large language models (LLMs) \citep{devlin2019bert,radford2018improving,radford2019language}. 
These trends have often been based around
architectural shifts,
to/from the ubiquity of recurrent neural networks (RNNs) \citep{hochreiter1997lstm}, 
and then into reliance on attention mechanisms \citep{bahdanau2015neural}
and the subsequent proliferation of self-attention leading to transformer-based LLMs~\citep{vaswani2017attention}.

\noindent\textbf{Optimization for pre-trained word embedding.} In the domain of pre-trained word embeddings, optimization methods are the essential that govern the performance and efficacy of the resulting word vectors. Early word embedding methods, like word2vec \citep{mikolov2013efficient} used gradient descent-based strategies to maximize context word likelihood, setting a foundation for subsequent models. Subsequent evolution led to the introduction of GloVe, which refined the optimization process by formulating a cost function based on global word co-occurrence statistics, merging local context and global matrix factorization methods to improve word representation. Despite its effectiveness, the performance of GloVe's optimization is limited by its predefined context window size to capture broader context \citep{pennington2014glove}. 

Following significant development in the optimization of pre-trained word embeddings was revealed by \citealp{NIPS2014_feab05aa}. They demonstrated that the skip-gram model with negative sampling implicitly executes matrix factorization on a word-context matrix which represents the pointwise mutual information (PMI) of the respective word-context pairs, emphasizing the critical role of matrix factorization in optimization techniques. This idea led to the understanding of how PMI-based word embeddings can encapsulate meaningful semantics \citep{arora-etal-2016-latent}. Recent study (\citealp{bojanowski2017enriching}) further improved performance with subword embeddings, treating each word as a bag of character n-grams, particularly benefiting morphologically rich languages. Current research even extends these techniques to sentence and paragraph levels for more efficient representations \citep{arora2017asimple}.

\noindent\textbf{Dimensionality Reduction with Embedding.} Advances in dimensionality reduction have significantly contributed to word embeddings. Traditional techniques, such as PCA \citep{Jolliffe:1986} and SVD \citep{klema1980singular}, transformed high-dimensional data into manageable lower-dimensional space, albeit with information loss. More recent works like \citealp{liu2016kernelized} introduced novel methods like Kernelized Matrix Factorization (KMF) that rejuvenated traditional matrix factorization techniques. Additionally, Heidenreich et al. \citep{heidenreich2022eigennoise} elucidated the deep connection between word representation algorithms and co-occurrence matrix factorization. The BERT model \citep{devlin2019bert}, despite its high-dimensionality, efficiently captures word semantics using dimensionality reduction techniques within a transformer architecture. 

However, the techniques talked about above always involve training neural networks. A method that combines dimensionality reduction techniques with leveraging co-occurrence statistics for learning efficient word representations, without the need for neural network training to learn explicit representations of tokens, would be beneficial and ideal. As our primary focus, it will be discussed in the following section.

\textbf{Language Model training and fine-tuning.} 
In the days following the advent of the Transformer model, when Large Language Models (LLMs) were not as prevalent as they are today, the predominant method for utilizing Language Models (LMs) was through a process of pre-training and subsequent fine-tuning for specific downstream tasks. With the model size not as large as today's, it is not as expansive as fine-tuning a LLM. Consequently, 1) fine-tuning a language model for a specific purpose was less computationally intensive, and 2) the intrinsic properties of fine-tuning ensured that models could always achieve better performance through task-specific fine-tuning. However, with training language models at scale becoming possible and the dominant paradigm of carrying out NLP research, the cost of doing LM-related experiments has also increased. Despite the extraordinary power and utility of LLMs, the training process usually takes days and costs a huge amount of money while sometimes finding it hard to outperform smaller, fine-tuned language models \citep{liu2022fewshot} on specific downstream tasks. In section \ref{sec: experiment}, we conduct experiments both for language model training and fine-tuning to demonstrate two useful scenarios to fit bit-cipher into the modern LLM world.


\section{Bit-Cipher}
\label{bit-cipher}


\subsection{Definition of Bit-cipher}
Standard-basis encoding is unavoidable for NLP applications, as one must always encode tokens from a given model's vocabulary: $W$. 
This makes dimensionality reduction \textit{necessary} for NLP applications, as the combinatorial overhead on the model parameters required to process $|W|$-dimensional hidden states becomes tremendous inside of models. While dimensionality reduction can be handled via gradient-based optimization in DL systems, the random nature of DL optimization obfuscates the meaning of low dimensions. However, we conjecture that a similar and explicit encoder-decoder-style factorization of standard-basis information exists.

Supposing each token $t$ in $W$ has identity modified from the usual one-hot vector as follows: (1) select a `low' dimension: $b\leq|W|$, and (2) assign a unique bit-vector, $\eta_{t}\in\{0,1\}^b$ to each. We base our approach on a distinguishability hypothesis: which expects that a `good' order for the bits distinguishes the highest-frequency tokens best, and has latitude to assign similar-frequency tokens similar vectors, meaning word vectors are assigned based on unigram frequency ranking. Working along these lines, we define $b$-bit encipherment as the process of assigning probabilistically normalized (e.g., vectors are probabilistic vectors and the modulus of vectors is 1) with all $b$-bit vectors in a `smooth' order, by inducting the order that $i=1$: assigns the set of $b$ standard basis vectors: $\mathcal{V}^b_1$ to the $b$ most-frequent tokens (generalizing one-hots/standard bases); $i=2$: adds standard-basis vectors to those from $\mathcal{V}^b_{k-1}$ in reverse order of assignment, while filtering for unique bit-vectors in $\{0,1\}^b$; $i=3$: repeats step $i=2$. $b$-bit vectors are then normalized for encipherment: $v_t = \eta_{t}/\|\eta_{t}\|_1$. 
\subsection{Modeling noise in observations}
To assure that co-occurrence matrices are dense, we modify the base representation of the model from sparse, one-hot, to dense vectors of the same size. We first form a model: $\beta\in (0,1)^N$, for the portion of time that each $i$-token's observations are (non-)erroneous as the definition shown above. Assuming that the highest-frequency tokens will be the least erroneously observed, we assume that only one error will be observed relative to each token's observed frequency, that is: $\beta_i = f_i/(f_i + 1)$, where $f_i$ is the unigram frequency of token \textit{i}. Next and regardless of the token that is observed, we wish to modify its one-hot vector according to the probabilities that any different, $j$-token, should have been observed, instead, which will take the form of another vector: $\sigma\in (0,1)^N$, but which is normalized: $\|\sigma\|_1 = 1$, and so define these other-token observation probabilities as: $\sigma_j = (1 - f_j/M)/(N - 1)$. To understand $\sigma$ intuitively, we first note that 1-minus each token's unigram probability: $1 - f_j/M$ expresses the probability of each token not being observed. Hence, the model $\sigma$ assumes that these (non-mutually exclusive) probabilities weight a distribution for the other token that should've been observed. For each one-hot vector, $y_i$, we then pull together these pieces to define noisy/dense vectors as: $\nu_i = \beta_iy_i + (1 - \beta_i)\sigma$, which form the embedding layers used in all language modeling architectures.

\subsection{Rundown of procedurally building cipher embeddings}
\label{sec:rundown}

Knowing the definition and the encoding method of noise into cipher embedding enables us the procedural generation of word vectors. With the given dimension $d$, the bit-cipher algorithm is capable of generating the number of  $2^d-1$ vectors. In details of how the process works: The procedure operates in two steps: Initially, a set of probabilistic vectors, referred to as "plain vectors", is generated in accordance with the given \textit{definition}.  Subsequently, noise information is encoded based on the analysis of the ratio of document frequency to word frequency, denoted as $r_i = d_i / f_i$.  This ratio determines the extent of noise information encoded into plain vectors. Specifically, words with high word frequency but low document frequency yield a small ratio, indicating that the word is noisy within the entire training set. Consequently, more noise information is "baked" into the plain vectors and vice versa. This is achieved using the formula: $\nu_i = \beta_iy_i + (1 - \beta_i)\sigma$ producing the final set of cipher embeddings  The pseudocode of how exactly the algorithm can be implemented is also shown in Fig\ref{alg:bit-cipher}

\subsection{Illustration through a concrete sample case — 5-bit cipher}

\begin{wrapfigure}{r}{0.5\textwidth}
   \vspace{-20pt} 
   \centering
   \includegraphics[width=0.43\textwidth]{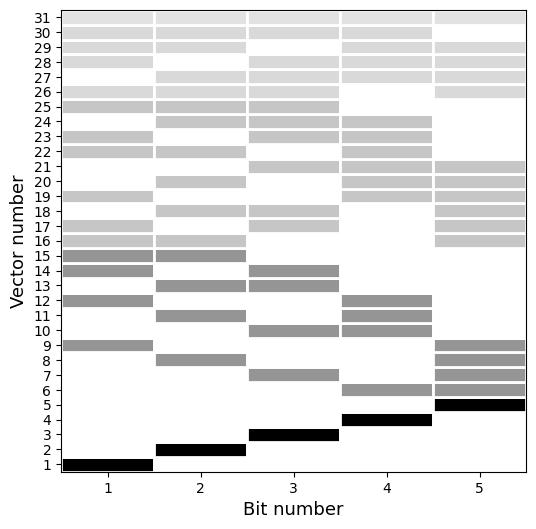}
   \caption{$5$-bit example, carried out over its largest vocabulary size of $2^5 - 1 = 31$ vectors (rows). }
   \label{fig:5-bit-example}
   \vspace{-20pt} 
\end{wrapfigure}

As depicted in Fig\ref{fig:5-bit-example}, an example with 5 bits is illustrated. In this scenario, the bit-cipher algorithm can produce $31$ distinct vectors with the capability of handling a corpus contain 31 unique tokens with each represented by a unique 5-bit vector. To elucidate the operation of the algorithm, consider the following steps visualized in the figure:\\
    1. The first vector corresponds to the most frequent word in the corpus, assigning the bit-number 1 a value of 1, and all others a value of 0. \\
    2. The second vector, representing the next frequent word, assigns bit-number 2 a value of 1, with all other bits set to 0. This pattern continues for the top  $C(5,1)=5$ words, assigning a value of 1 to the corresponding position based on ranking.\\
    3. Upon reaching the count of $5$, words ranked between $[C(5,1)+1, C(5,1)+C(5,2)]$ e.g., $[6,15]$ are assigned values in reverse order of index; two positions are assigned a value of $1 / 2 = 0.5$, and all others are 0.\\
    4. For words ranked within the intervals of $[C(5,2)+1,C(5,2)+C(5,3)]$, $[C(5,3)+1,C(5,3)+C(5,4)]$, and $[C(5,4)+1,C(5,4)+C(5,5)]$, values are assigned to positions following the same logic.


Finally, each unique token is allocated a unique vector. By incorporating noise information relative to the distribution of words across various documents, the finalized version of bit-cipher embedding is obtained.

\begin{figure*}[t]
\begin{algorithmic}[1]
\Procedure{Bit-Cipher}{$N,b$}\Comment{Construct a $b$-bit cipher of $N\leq 2^{b-1}$ dimensions.}
    
    \State $B^{(0)} \gets [\vec{0}]$
    \For{$k = 1, \cdots, b$} \Comment{\textbf{1.} Initialize sets for differently-normed bit-vectors.}
        \State $B^{(k)} \gets []$
    \EndFor
    \State $U, V\gets \{0\}^{N\times b}, \{0\}^{N\times b}$ 
    \State $i,j,k \gets 0,0,1$
    \For{$n=1, \cdots, N$} 
        \While{$V_n = \vec{0}$}\Comment{\textbf{2.} Find the next norm-$k$ (or $k+1$) bit-vector.}
            \State $u \gets {\rm Abs}\left(B^{(k-1)}_{j} - I_i\right)$ 
            \If{$\|u\|_1 = k$ and $u\notin B_k$}\Comment{\textbf{3.} The norm must be $k$ and the vector unused.}
                \State $B^{(k)} \gets {\rm Concatenate}\left(B^{(k)}, [u]\right)$ 
                \State $V_n \gets u/\|u\|_1$\Comment{\textbf{4.} Norm the bit-vector and assign it.}
                \State $U_n \gets u$ 
                
            \EndIf
            \State $j \gets j + 1$ 
            \If{$j = |B^{(k-1)}|$}\Comment{\textbf{5.} Change basis vector/component of modification.}
                \State $j \gets 0$ 
                \State $i \gets i + 1$ 
                \If{$i = b$} \Comment{\textbf{6.} Reverse the $k$-bit vector order and increment $k$.}
                    \If{$k = 1$}
                        \State $I \gets {\rm Reverse}\left(I\right)$
                    \EndIf
                    \State $i \gets 0$ 
                    \State $B^{(k)} \gets {\rm Reverse}\left(B^{(k)}\right)$
                    \State $k \gets k + 1$
                \EndIf
            \EndIf
        \EndWhile
    \EndFor 
    \State \textbf{return} $U, V$\Comment{\textbf{7.} Return matrices for deciphering and enciphering.}
\EndProcedure
\end{algorithmic}
\caption{Bit-Cipher algorithm. After 1) initialization, the algorithm must 2) find new bit-vectors in decreasing order of discernability, by 3) identifying bit-vectors of increasing norm (that have not yet been assigned) via translations of $k-1$-bit vectors by standard basis vectors. Unassigned bit-vectors are then 4) normed for encipherment and assigned, along with the raw bit-vectors, which can be used for deciphering $b$-dimensional predictions. Whenever the collection of $k-1$-bit vectors no longer has any unassigned $i$-component modifications, 5) the basis vector/component of modification must be incremented, and when this is the case for all last-component modifications, it's determined that there are no unassigned $k$-bit vectors, necessitating a 6) reversal of the $k$-bit vector order, which maintains smoothe transitions of discernability, upon future assignment. 7) Once all $N$ dimensions have been assigned a bit-vector (and normed counterpart), the matrices containing these vectors are returned.}\label{alg:bit-cipher}
\end{figure*}

\subsection{Bit-Cipher training detail}
\label{sec:train}
To illustrate the efficacy of cipher embeddings, models were trained on the CommonCrawl dataset using \textbf{Cat} (concatenation) and \textbf{Sum} (summation) methods for aggregating contextual information, informed by the \textit{bits} hyperparameter, \textit{log=True}, and \textit{dtype='df'}. The latter two parameters enhanced sensitivity to infrequent words and adjusted noise levels based on word's document frequency, optimizing focus on distinctive words and mitigating biases.

The bit-cipher models are trained on five different scales of data-size: 0.5B-token, 1B-token, 2B-token, 4B-token, and 8B-token with setting up different radius (window size) and bits. Through incremental increase of the data-size, we aim to understand how model performance adjusts with the intake of more data. Although this data-size range is relatively small compared to other pre-trained word embedding methods, like GloVe trained with 42B and 840B tokens \citep{pennington-etal-2014-glove} and word2vec trained on Google News with 100B tokens \citep{mikolov2013efficient}. All that is being said here is to validate the efficiency of bit-cipher as a means of learning representation, which we further corroborate through a series of probing experiments in the section.\ref{sec:probing}.

Standard spaCy tokenization \citep{spacy} was used for preprocessing, and models underwent a two-step training procedure as per sec \ref{sec:rundown} \& Figure \ref{alg:bit-cipher}. Contextual information was integrated using \textbf{Cat} or \textbf{Sum} methods, with \textbf{Cat} models achieving representation lengths of 200d to 1600d (following the exponent of 2 times 100) across different bit settings and \textbf{Sum} models blending context information through element-wise addition, yielding a total of 60 models across varied radii and data sizes.



For comparability across models, we derived word embeddings from the GlOVe 6B embeddings, encompassing 400,000 tokens and with tokens appear in the evaluation datasets, yielding a total of 419,374 unique word embeddings. Any words identified within the context window that did not exist in our curated word-list were labeled as out-of-vocabulary (OOV) and were consequently assigned a distinctive embedding. This strategy for managing OOV words contributes to memory optimization, given that it mandates the processing of only a particular subset of words.

\section{Probing experiments for linguistic features capture}
\label{sec:probing}

\subsection{Probing models}
The conduction of Probing experiments are inspired by \citep{hewitt2019designing} with designing  POS tagging with the Georgetown University Multilayer (GUM) dataset \citep{Zeldes2017}, Named Entity Recognition (NER) using CoNLL-2003 shared benchmark dataset \citep{tjong-kim-sang-de-meulder-2003-introduction} to evaluate the performance of bit-cipher.  

\noindent\textbf{Named Entity Recognition.} NER probing experiment is conducted by CoNLL-2003 shared benchmark dataset which is a collection of data about Reuters newswire articles containing four different entity types: persons (PER), organizations (ORG), locations (LOC) and miscellaneous names (MISC). The probing model for NER is trained on CoNLL-2003 training data using CoNLL-2003 validation set for hyperparameter tuning. We follow the simplest and most straightforward setup with training an MLP by only using the bit-cipher embedding as the feature and directly adopt labels in the CoNLL-2003 dataset using the label-to-index method to convert each label into a unique number to setup the input and output of the probing model. 

\noindent\textbf{Part-of-speech (POS) tagging.} Part-of-speech tagging is a task of assigning labels to each word with its corresponding grammatical category, such as noun, verb, adjective, etc. The Georgetown University Multilayer (GUM) dataset is a richly annotated corpus that contains comprehensive linguistic features.  We extract the POS tagger of words in the GUM and train an MLP following the same setup as the NER experiment using the bit-cipher embeddings as the input and POS taggers as output.

\subsection{ probing model building details}

After obtaining the bit-cipher embeddings following \ref{sec:train}, we applied a two-step post-processing to refine the word representations, and all probing experiments used this refined version of bit-cipher. Initially, a whitening transformation was employed to eliminate redundancies and normalize the embeddings, ensuring linearly uncorrelated word vectors with uniform variance, reducing inherent bias and making the distribution of embeddings more consistent \citep{kessy2016optimal}.

Next, we implemented mean-centering and L2 Normalization on each vector to address shifts in statistical distribution, inherent in probabilistic vectors like bit-cipher, which could cause inconsistencies in magnitude. This process stabilized the numerical representations, making them robust, and ensuring unbiased and scale-independent comparisons between word vectors.

For probing experiments, a 2-layer Multi-Layer Perception (MLP) was utilized, incorporating LeakReLU activation to mitigate the vanishing gradient problem, and a dropout rate of 0.5 for regularization \citep{xu2015empirical}. The output layer featured a LogSoftmax function, maintaining numerical stability and a balanced probability distribution, key for optimal performance.

\subsection{Probing experiments results}

Probing experiments conducted on 100 separate bit-cipher embedding sets are presented in \textbf{Tables \ref{tab:sum-pos}--\ref{tab:cip-ner}}. Their results at POS tagging and NER demonstrate noticeable and perhaps expected variations in performance. We see clearly that cipher-only models generally don't improve with increases of data (\textbf{Tabs. \ref{tab:cip-pos},\ref{tab:cip-ner}}), which is sensible given that ciphers only require ranking information and word frequency ranks converge over relatively little data. 

\begin{wraptable}{r}{0.5\textwidth}

\vspace{-10pt} 
\caption{Comparison of a $300$-dimensional word2vec model against $200$-dimensional models (all others) on probing experiments. Note: both of GloVe and word2vec were pre-trained externally using a larger radius of $10$, by comparison to the Sum and Cat models presented, which were trained using $r = 4$.(values in the table are shown as accuracy with F1-score in Parentheses)}
\centering
\renewcommand{\arraystretch}{1.5}
\begin{tabular}{lcc}
\hline
\textbf{Models} & \textbf{POS} & \textbf{NER} \\
\hline
word2vec & 81.20 (80.80) & 78.55 (77.44) \\
GloVe.6B & 85.50 (86.09) & \textbf{91.70 (92.18)} \\
Cipher  & 75.23 (73.58) & 86.19 (84.17) \\ 
Cipher (Sum) & 85.67 (86.04) & 90.67 (91.32) \\ 
Cipher (Cat) & \textbf{86.05 (86.32)} & 90.96 (91.51) \\\hline
\end{tabular}
\label{tab:model-comparison}
\end{wraptable}

\begin{figure}[t]
    \centering
    
    \begin{subfigure}[b]{0.48\textwidth}
        \includegraphics[width=\textwidth]{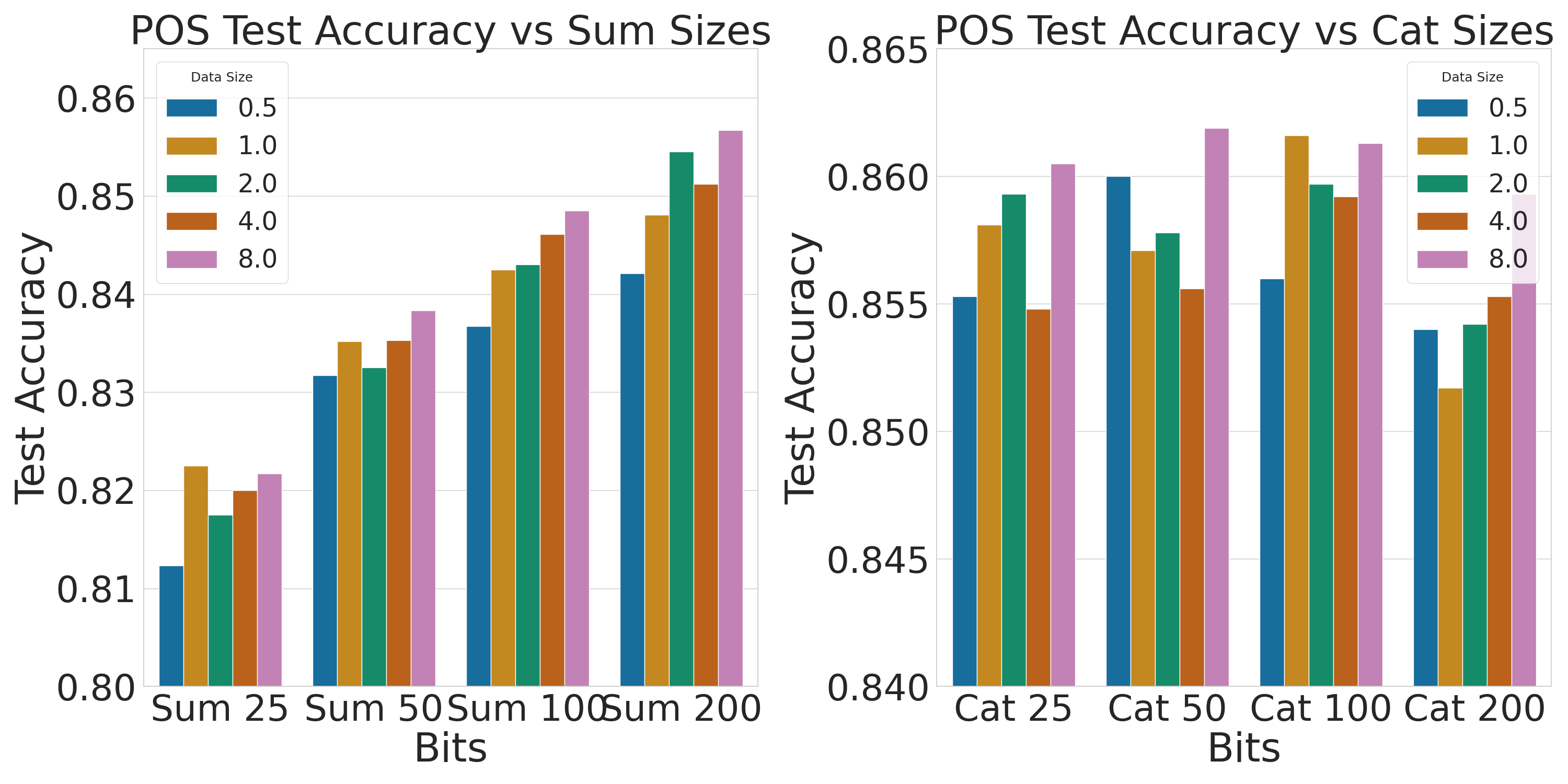}
        \caption{Comparison between how the performance of the \textbf{Cat} and \textbf{Sum} models in the POS experiment changes with the change in data size and Bits with setting radius = 4}
        \label{fig:pos-comparison}
    \end{subfigure}
    \hfill
    \begin{subfigure}[b]{0.48\textwidth}
        \includegraphics[width=\textwidth]{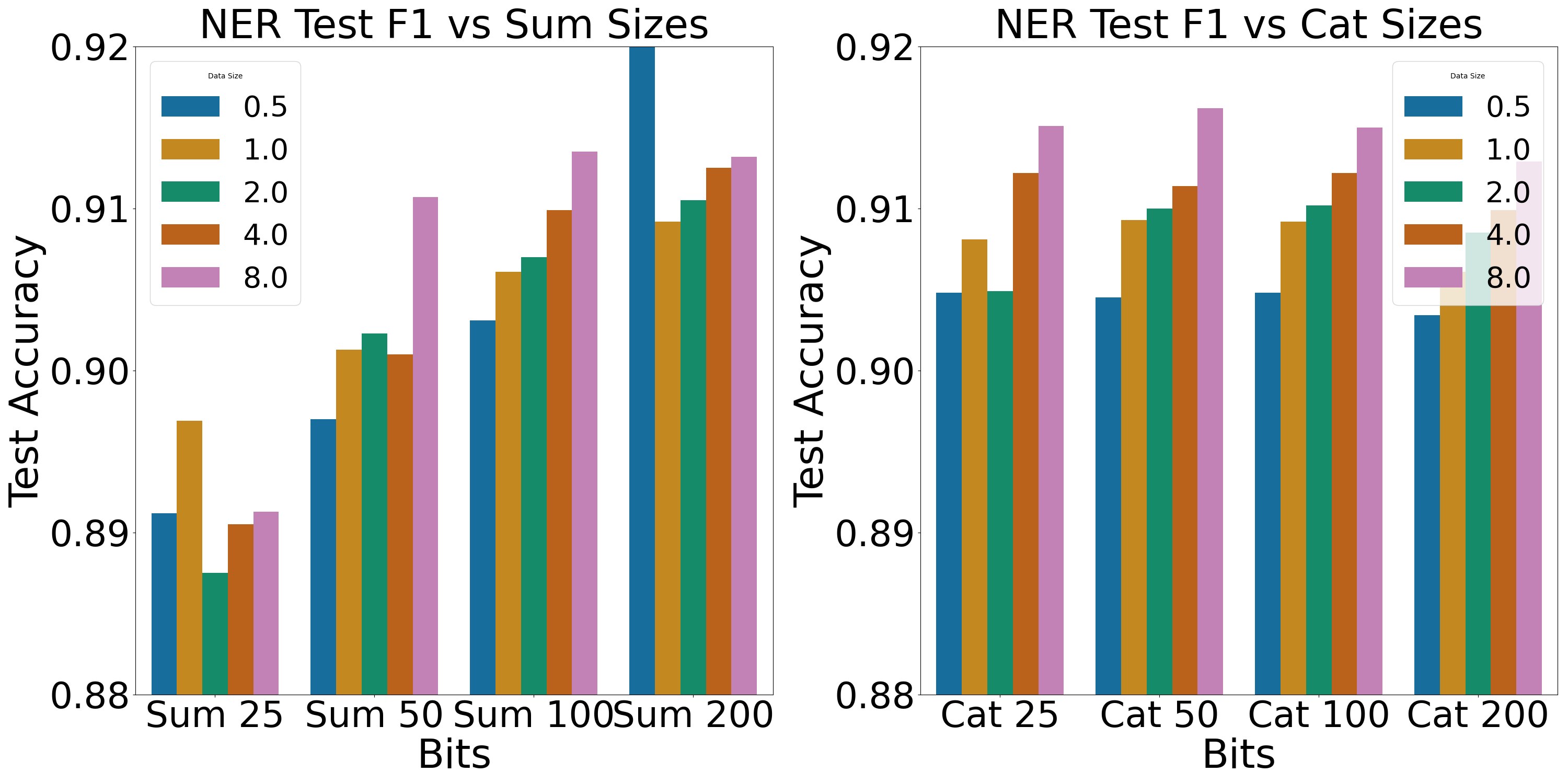}
        \caption{Comparison between how the performance of the \textbf{Cat} and \textbf{Sum} models in the NER experiment changes with the change in data size and Bits with setting radius = 4}
        \label{fig:ner-comparison}
    \end{subfigure}
    
    \caption{Comparison of \textbf{Cat} and \textbf{Sum} models in POS and NER experiments}
    \label{fig:comparison}
\end{figure}

For both \textbf{Sum} and \textbf{Cat} models (\textbf{Tabs. \ref{tab:sum-pos}--\ref{tab:cat-ner}}), we see marked improvements from training over increased volumes of data, as observed from \textbf{Figure \ref{fig:pos-comparison} \& \ref{fig:ner-comparison}} that when the \textit{bits} is fixed, increasing the data size often results in improved model performance. Furthermore, in the case of Sum models, a clear performance gain is observed with an increase in the value of bits with bits = 200 set of models consistently demonstrate the highest performance. However, the performance is likewise sensible, assuming that the quadratic co-frequency information in co-occurrences requires more data to stabilize. However, \textit{between} the \textbf{Sum} and \textbf{Cat} models, we note that \textbf{Cat} models improve over increases in data with greater \textit{stability}---they scale more reliably as shown in \textbf{Figure \ref{fig:pos-comparison} \& \ref{fig:ner-comparison}} that with fixing bits, 8B models always have the best performance. Moreover, we find that \textbf{Cat} models appear to consistently outperform same-dimension \textbf{Sum} models, despite being constrained to fewer bits, as can be seen from the cross-section of comparable models presented in \textbf{Tab.~\ref{tab:model-comparison}}. The inconsisency of model performance tendency is partially due to the fact that we did not do any preprocessing of the data except lowercase when training the bit-cipher. With refined preprocessing, the information gain would be even obviously to observe with the increase of data size.

When similar quantities of data are utilized, models that are more performant than word2vec, as well as quite comparable to GloVe, can be trained from bit-cipher co-occurrences. This can be see directly in \textbf{Tab.~\ref{tab:model-comparison}} for 200-dimensional bit-cipher models, which we compare to an externally-trained 300-dimensional word2vec model, a 200-dimensional GloVe and bit-cipher models. On its own, the noised cipher out-competes word2vec, while relatively-low radius ($r=4$) \textbf{Sum} and \textbf{Cat} models perform comparably to a set of GloVe embedding, which were also externally trained. Despite the \textbf{Sum} and \textbf{Cat} models both utilizing a substantially smaller radius ($r=4$) than GloVe ($r=10$), we see that both of the comparable co-occurrent bit-cipher models out-perform GloVe at POS tagging, and perform comparably at NER. Finally, we note that these results \textit{rank the bit-cipher at position 20}
amongst the NER models retained on a well-known/public page: Tracking Progress in Natural Language Processing~\citep{ruder2023progress}, and moreover, present POS tagging results quite similar to other strong baselines~\citep{ruder-plank-2018-strong}, whose model architectures tend to be much more complex and expressive than the MLPs used in our probing experiments.




\section{Experiments with language models}
\label{sec: experiment}

Despite the models we’ve trained to exhibit the efficiency of cipher, and the probing experiments conducted to demonstrate its competitiveness with classic pre-trained word embeddings, we further explore what we believe to be two of the bit-cipher's most valuable applications — LM training and efficient LM fine-tuning.

\label{finetune}

\subsection{A showcase of building language models with cipher}

\begin{wrapfigure}{r}{0.5\textwidth}
   \vspace{-15pt} 
   \centering
    \includegraphics[width=0.48\textwidth]{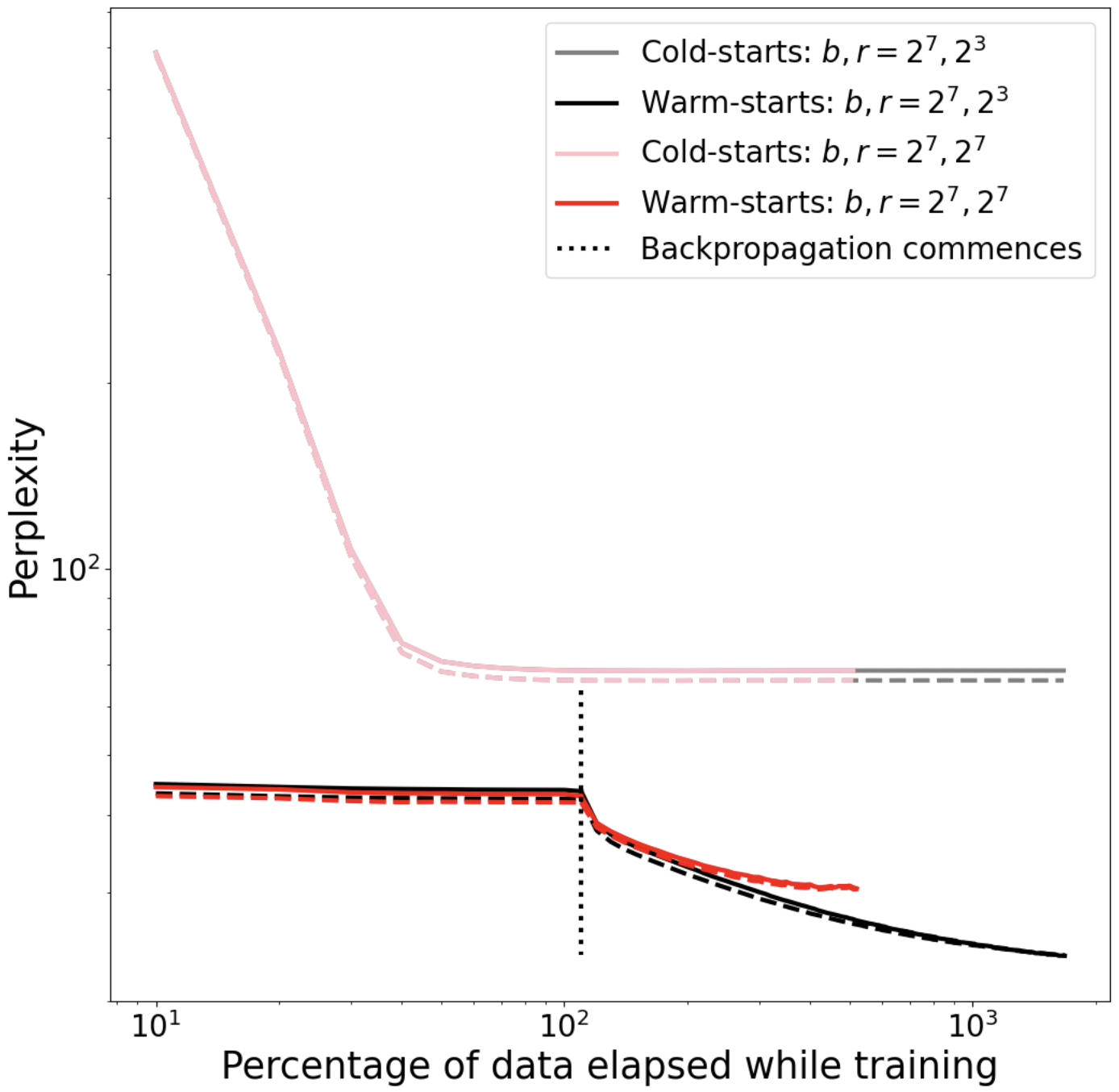}
    \caption{cold-start vs. warm-start perplexity curve with training processing }
   \label{fig:cipher_lm}
   \vspace{-10pt} 
\end{wrapfigure}

Firstly, the potential application of cipher is in the efficient training of language models. By using bit-cipher to construct the embedding layer of the language model and integrating it into the model's training process, we could potentially improve training efficiency and reduce the demand for computational resources.

Models are trained from scratch, utilizing both cold-start and warm-start approaches, with standard transformers \citep{vaswani2017attention}. Our approach involves initially training bit-cipher with the BabyLM 10M dataset and replacing the randomly initialized embeddings in the warm-start model. An additional technique employed in the warm-start cipher with language model training involves freezing the embedding layer before the model is trained and subsequently unfreezing it for further optimization using backpropagation. This freezing/thawing technique offers two benefits: (1) As the embedding layer is the first layer in any language model, it typically requires the most optimization time through backpropagation and is thus the most expensive layer. By initially freezing this layer, (2) we avoid the deterioration of model performance, in terms of perplexity, that can occur when the sensitive and delicate embeddings are modified during warm-start training. Therefore, the warm-start model adopts a two-step training procedure: initially freezing the embedding layer and proceeding with regular training, followed by unfreezing the embedding layer for further optimization through backpropagation. Cold-start models adhere to the traditional training approach, initializing all parameters randomly and optimizing them through backpropagation.

We conducted experiments using two sets of cipher embeddings: one with bits=$2^7$ and radius=$2^7$, and another with bits=$2^7$ and radius=$2^3$. The comparison of perplexity between warm-start and cold-start models is illustrated in \textbf{Fig}\ref{fig:cipher_lm}. The figure distinctly demonstrates that not only do warm-start models begin with a superior start, but they can also be further optimized through backpropagation, making this an overall more effective method for training language models.

\subsection{Language model fine-tuning with bit-cipher}

In addition to using bit-cipher as part of LM training, we find it's also promising to use the algorithm for efficient LM fine-tuning. The traditional finetuning process, which necessitates retraining the model on a task-specific dataset, remains costly, leading to the exploration of zero-shot, few-shot, and in-context learning strategies, prioritizing performance and efficiency. Although these methods effectively extract useful features learned during training, there is a known trade-off; for instance, prompted models do not always outperform fine-tuned models. Fine-tuned models, trained for a specific purpose on one or a series of related datasets for a downstream task, typically achieve state-of-the-art (SOTA) results.

A paradigm of fine-tuning that balances performance and training efficiency is desirable, allowing for the deployment of numerous specific-purpose models with superior performance that are less costly than training large models. In our method, we first train cipher embeddings on the fine-tuned dataset to acquire what we term “cipher fine-tune embeddings”, then replace the embedding layer in the pre-trained language models with these cipher-fine-tuned embeddings, designed with specific fine-tune objectives. The efficiency of cipher training renders this step cost-effective, enhancing overall model efficiency.

\begin{figure}
    \centering
    
    \begin{subfigure}[b]{0.32\textwidth}
        \includegraphics[width=\textwidth]{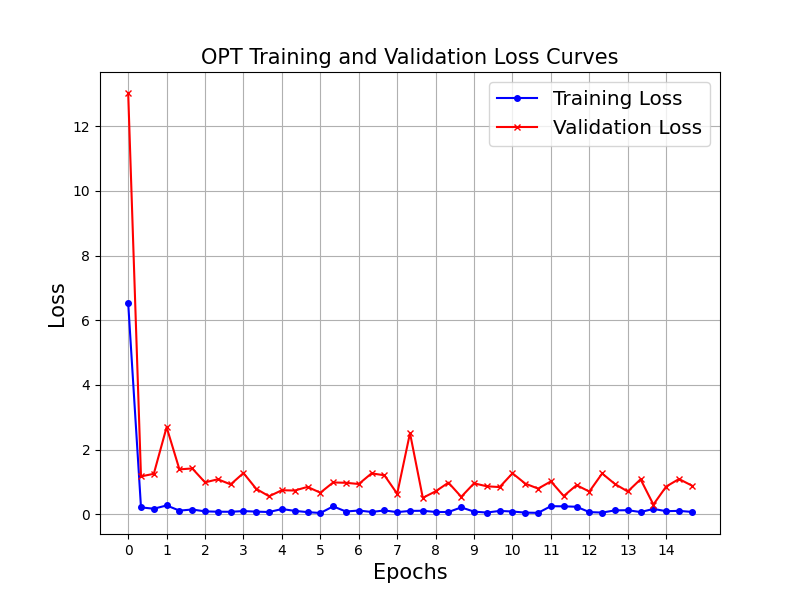}
        \caption{train/dev losses of OPT}
        \label{fig:opt_loss}
    \end{subfigure}
    \hfill 
    \begin{subfigure}[b]{0.32\textwidth}
        \includegraphics[width=\textwidth]{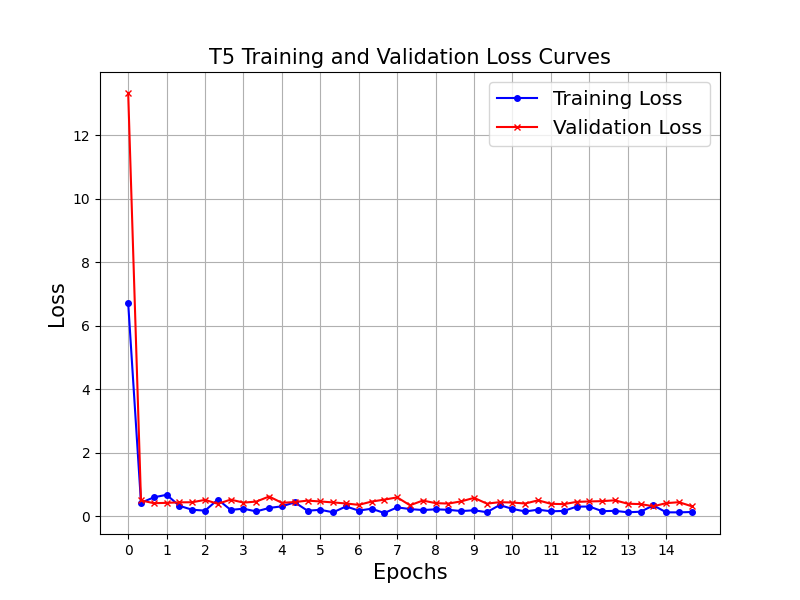}
        \caption{train/dev losses of T5}
        \label{fig:t5_loss}
    \end{subfigure}
    \hfill
    \begin{subfigure}[b]{0.32\textwidth}
        \includegraphics[width=\textwidth]{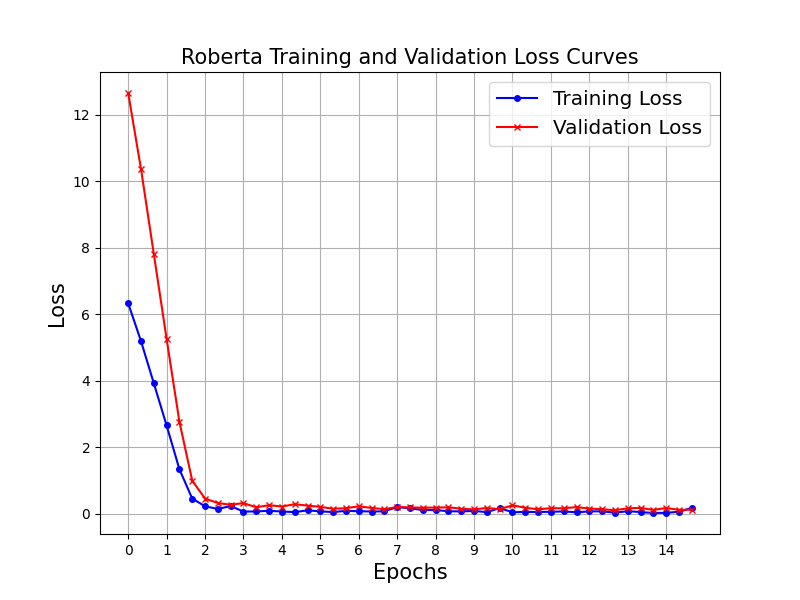}
        \caption{train/dev losses of Roberta}
        \label{fig:roberta_loss}
    \end{subfigure}
    
    \caption{Loss curves of different models}
    \label{fig:loss_curves}
\end{figure}

We selected three language models: T5, Roberta, and OPT, and fine-tuned them on the 10M dataset provided by BabyLM \citep{warstadt2023papers}. \citep{eval-harness} are conducted following a two-step process: (1) Train cipher embeddings with the dataset used for the specific fine-tuning purpose. (2) Replace the embedding layer of the language model designated for fine-tuning with cipher embeddings. This approach enables models to converge more rapidly compared to traditional methods, as illustrated by three training/dev curves\ref{fig:loss_curves}, showcasing the speed of fine-tuning that fine-tuned model can quickly converge to low-enough training and developing losses which result in the acceleration of fine-tuning process as well as the reduction of computational costs.












\section{Conclusion}

In conclusion, in this paper, we introduce Bit-cipher, a novel and efficient method of learning word representations. By using this strategy, we acquire static pre-trained embeddings controlled by dimensionalities set with \textit{bits} and learn contextual information by simple vector addition, eliminating the need for neural network training.  Consequently, the model learns explicit statistical information from large text with strong interpretability. Our results show that \textbf{Cat} models consistently outperform \textbf{Sum} models across different dimensions and data sizes, demonstrating greater stability and superior performance even when constrained to fewer bits. However, \textbf{Sum} models also show merit, especially with the potential for further architectural improvements. Furthermore, by comparing with GloVe and word2vec, the competitive performance of bit-cipher is further validated. Additionally, language modeling experiments are conducted through both showing the efficacy of cipher as part of language model training and an efficient alternative to the traditional fine-tuning process. Overall, we see the bit-cipher as an efficient and high-performing alternative to classic pre-trained word embedding methods, with significantly reduced costs, offering a unique niche in the LLM era based on efficiency and interpretability---without performance compromise.
\clearpage


\bibliography{iclr2024_conference}

\begin{thebibliography}{36}
\providecommand{\natexlab}[1]{#1}
\providecommand{\url}[1]{\texttt{#1}}
\expandafter\ifx\csname urlstyle\endcsname\relax
  \providecommand{\doi}[1]{doi: #1}\else
  \providecommand{\doi}{doi: \begingroup \urlstyle{rm}\Url}\fi

\bibitem[Arora et~al.(2016)Arora, Li, Liang, Ma, and Risteski]{arora-etal-2016-latent}
Sanjeev Arora, Yuanzhi Li, Yingyu Liang, Tengyu Ma, and Andrej Risteski.
\newblock A latent variable model approach to {PMI}-based word embeddings.
\newblock \emph{Transactions of the Association for Computational Linguistics}, 4:\penalty0 385--399, 2016.
\newblock \doi{10.1162/tacl_a_00106}.
\newblock URL \url{https://aclanthology.org/Q16-1028}.

\bibitem[Arora et~al.(2017)Arora, Liang, and Ma]{arora2017asimple}
Sanjeev Arora, Yingyu Liang, and Tengyu Ma.
\newblock A simple but tough-to-beat baseline for sentence embeddings.
\newblock 2017.

\bibitem[Bahdanau et~al.(2015)Bahdanau, Cho, and Bengio]{bahdanau2015neural}
Dzmitry Bahdanau, Kyung~Hyun Cho, and Yoshua Bengio.
\newblock Neural machine translation by jointly learning to align and translate.
\newblock In \emph{3rd International Conference on Learning Representations, ICLR 2015}, 2015.

\bibitem[Bojanowski et~al.(2017{\natexlab{a}})Bojanowski, Grave, Joulin, and Mikolov]{bojanowski2017enriching}
Piotr Bojanowski, Edouard Grave, Armand Joulin, and Tomas Mikolov.
\newblock Enriching word vectors with subword information.
\newblock \emph{Transactions of the Association for Computational Linguistics}, 5:\penalty0 135--146, 2017{\natexlab{a}}.
\newblock ISSN 2307-387X.

\bibitem[Bojanowski et~al.(2017{\natexlab{b}})Bojanowski, Grave, Joulin, and Mikolov]{bojanowski2017tricks}
Piotr Bojanowski, Edouard Grave, Armand Joulin, and Tomas Mikolov.
\newblock Bag of tricks for efficient text classification.
\newblock In \emph{Proceedings of the 15th Conference of the European Chapter of the Association for Computational Linguistics: Volume 2, Short Papers}, pp.\  427--431. Association for Computational Linguistics, 2017{\natexlab{b}}.

\bibitem[Devlin et~al.(2019)Devlin, Chang, Lee, and Toutanova]{devlin2019bert}
Jacob Devlin, Ming-Wei Chang, Kenton Lee, and Kristina Toutanova.
\newblock Bert: Pre-training of deep bidirectional transformers for language understanding, 2019.

\bibitem[Gao et~al.(2021)Gao, Tow, Biderman, Black, DiPofi, Foster, Golding, Hsu, McDonell, Muennighoff, Phang, Reynolds, Tang, Thite, Wang, Wang, and Zou]{eval-harness}
Leo Gao, Jonathan Tow, Stella Biderman, Sid Black, Anthony DiPofi, Charles Foster, Laurence Golding, Jeffrey Hsu, Kyle McDonell, Niklas Muennighoff, Jason Phang, Laria Reynolds, Eric Tang, Anish Thite, Ben Wang, Kevin Wang, and Andy Zou.
\newblock A framework for few-shot language model evaluation.
\newblock September 2021.
\newblock \doi{10.5281/zenodo.5371628}.
\newblock URL \url{https://doi.org/10.5281/zenodo.5371628}.

\bibitem[Heidenreich \& Williams(2022)Heidenreich and Williams]{heidenreich2022eigennoise}
Hunter~Scott Heidenreich and Jake~Ryland Williams.
\newblock Eigennoise: A contrastive prior to warm-start representations, 2022.

\bibitem[Hewitt \& Liang(2019)Hewitt and Liang]{hewitt2019designing}
John Hewitt and Percy Liang.
\newblock Designing and interpreting probes with control tasks, 2019.

\bibitem[Hochreiter \& Schmidhuber(1997)Hochreiter and Schmidhuber]{hochreiter1997lstm}
Sepp Hochreiter and J{\"u}rgen Schmidhuber.
\newblock Lstm can solve hard long time lag problems.
\newblock \emph{Advances in neural information processing systems}, pp.\  473--479, 1997.

\bibitem[Honnibal et~al.(2020)Honnibal, Montani, Van~Landeghem, and Boyd]{spacy}
Matthew Honnibal, Ines Montani, Sofie Van~Landeghem, and Adriane Boyd.
\newblock "spacy: Industrial-strength natural language processing in python".
\newblock 2020.

\bibitem[Howard \& Ruder(2018)Howard and Ruder]{howard2018universal}
Jeremy Howard and Sebastian Ruder.
\newblock Universal language model fine-tuning for text classification.
\newblock In \emph{Proceedings of the 56th Annual Meeting of the Association for Computational Linguistics (Volume 1: Long Papers)}, pp.\  328--339, 2018.

\bibitem[Jolliffe(1986)]{Jolliffe:1986}
I.T. Jolliffe.
\newblock \emph{Principal Component Analysis}.
\newblock Springer Verlag, 1986.

\bibitem[Kessy et~al.(2016)Kessy, Lewin, and Strimmer]{kessy2016optimal}
Agnan Kessy, Alex Lewin, and Korbinian Strimmer.
\newblock {Optimal whitening and decorrelation}.
\newblock \emph{The American Statistician}, pp.\  1--6, December 2016.
\newblock ISSN 0003-1305.
\newblock \doi{10.1080/00031305.2016.1277159}.
\newblock URL \url{http://dx.doi.org/10.1080/00031305.2016.1277159}.

\bibitem[Klema \& Laub(1980)Klema and Laub]{klema1980singular}
V.~Klema and A.~Laub.
\newblock {The singular value decomposition: Its computation and some applications}.
\newblock \emph{IEEE Transactions on Automatic Control}, 25\penalty0 (2):\penalty0 164--176, 1980.

\bibitem[Levy \& Goldberg(2014)Levy and Goldberg]{NIPS2014_feab05aa}
Omer Levy and Yoav Goldberg.
\newblock Neural word embedding as implicit matrix factorization.
\newblock In Z.~Ghahramani, M.~Welling, C.~Cortes, N.~Lawrence, and K.Q. Weinberger (eds.), \emph{Advances in Neural Information Processing Systems}, volume~27. Curran Associates, Inc., 2014.
\newblock URL \url{https://proceedings.neurips.cc/paper_files/paper/2014/file/feab05aa91085b7a8012516bc3533958-Paper.pdf}.

\bibitem[Liu et~al.(2022)Liu, Tam, Muqeeth, Mohta, Huang, Bansal, and Raffel]{liu2022fewshot}
Haokun Liu, Derek Tam, Mohammed Muqeeth, Jay Mohta, Tenghao Huang, Mohit Bansal, and Colin Raffel.
\newblock Few-shot parameter-efficient fine-tuning is better and cheaper than in-context learning, 2022.

\bibitem[Liu et~al.(2016)Liu, Aggarwal, Li, Kong, Sun, and Sathe]{liu2016kernelized}
Xinyue Liu, Chara Aggarwal, Yu-Feng Li, Xiaugnan Kong, Xinyuan Sun, and Saket Sathe.
\newblock Kernelized matrix factorization for collaborative filtering.
\newblock In \emph{Proceedings of the 2016 SIAM International Conference on Data Mining}, pp.\  378--386. SIAM, 2016.

\bibitem[Mikolov et~al.(2013{\natexlab{a}})Mikolov, Chen, Corrado, and Dean]{mikolov2013efficient}
Tomas Mikolov, Kai Chen, Greg Corrado, and Jeffrey Dean.
\newblock Efficient estimation of word representations in vector space, 2013{\natexlab{a}}.

\bibitem[Mikolov et~al.(2013{\natexlab{b}})Mikolov, Sutskever, Chen, Corrado, and Dean]{mikolov2013distributed}
Tomas Mikolov, Ilya Sutskever, Kai Chen, Greg Corrado, and Jeffrey Dean.
\newblock Distributed representations of words and phrases and their compositionality.
\newblock In \emph{Proceedings of the 26th International Conference on Neural Information Processing Systems-Volume 2}, pp.\  3111--3119, 2013{\natexlab{b}}.

\bibitem[Panahi et~al.(2020)Panahi, Saeedi, and Arodz]{panahi2020word2ket}
Aliakbar Panahi, Seyran Saeedi, and Tom Arodz.
\newblock word2ket: Space-efficient word embeddings inspired by quantum entanglement, 2020.

\bibitem[Pennington et~al.(2014{\natexlab{a}})Pennington, Socher, and Manning]{pennington-etal-2014-glove}
Jeffrey Pennington, Richard Socher, and Christopher Manning.
\newblock {G}lo{V}e: Global vectors for word representation.
\newblock In \emph{Proceedings of the 2014 Conference on Empirical Methods in Natural Language Processing ({EMNLP})}, pp.\  1532--1543, Doha, Qatar, October 2014{\natexlab{a}}. Association for Computational Linguistics.
\newblock \doi{10.3115/v1/D14-1162}.
\newblock URL \url{https://aclanthology.org/D14-1162}.

\bibitem[Pennington et~al.(2014{\natexlab{b}})Pennington, Socher, and Manning]{pennington2014glove}
Jeffrey Pennington, Richard Socher, and Christopher~D Manning.
\newblock Glove: Global vectors for word representation.
\newblock In \emph{Proceedings of the 2014 conference on empirical methods in natural language processing (EMNLP)}, pp.\  1532--1543, 2014{\natexlab{b}}.

\bibitem[Peters et~al.(2018{\natexlab{a}})Peters, Neumann, Iyyer, Gardner, Clark, Lee, and Zettlemoyer]{peters-etal-2018-deep}
Matthew~E. Peters, Mark Neumann, Mohit Iyyer, Matt Gardner, Christopher Clark, Kenton Lee, and Luke Zettlemoyer.
\newblock Deep contextualized word representations.
\newblock In \emph{Proceedings of the 2018 Conference of the North {A}merican Chapter of the Association for Computational Linguistics: Human Language Technologies, Volume 1 (Long Papers)}, pp.\  2227--2237, New Orleans, Louisiana, June 2018{\natexlab{a}}. Association for Computational Linguistics.
\newblock \doi{10.18653/v1/N18-1202}.
\newblock URL \url{https://aclanthology.org/N18-1202}.

\bibitem[Peters et~al.(2018{\natexlab{b}})Peters, Neumann, Iyyer, Gardner, Clark, Lee, and Zettlemoyer]{peters2018deep}
Matthew~E. Peters, Mark Neumann, Mohit Iyyer, Matt Gardner, Christopher Clark, Kenton Lee, and Luke Zettlemoyer.
\newblock Deep contextualized word representations, 2018{\natexlab{b}}.

\bibitem[Radford et~al.(2018)Radford, Narasimhan, Salimans, and Sutskever]{radford2018improving}
Alec Radford, Karthik Narasimhan, Tim Salimans, and Ilya Sutskever.
\newblock Improving language understanding by generative pre-training.
\newblock 2018.

\bibitem[Radford et~al.(2019)Radford, Wu, Child, Luan, Amodei, and Sutskever]{radford2019language}
Alec Radford, Jeffrey Wu, Rewon Child, David Luan, Dario Amodei, and Ilya Sutskever.
\newblock Language models are unsupervised multitask learners.
\newblock \emph{OpenAI blog}, 1\penalty0 (8):\penalty0 9, 2019.

\bibitem[Rae et~al.(2022)Rae, Borgeaud, Cai, Millican, Hoffmann, Song, Aslanides, Henderson, Ring, Young, Rutherford, Hennigan, Menick, Cassirer, Powell, van~den Driessche, Hendricks, Rauh, Huang, Glaese, Welbl, Dathathri, Huang, Uesato, Mellor, Higgins, Creswell, McAleese, Wu, Elsen, Jayakumar, Buchatskaya, Budden, Sutherland, Simonyan, Paganini, Sifre, Martens, Li, Kuncoro, Nematzadeh, Gribovskaya, Donato, Lazaridou, Mensch, Lespiau, Tsimpoukelli, Grigorev, Fritz, Sottiaux, Pajarskas, Pohlen, Gong, Toyama, de~Masson~d'Autume, Li, Terzi, Mikulik, Babuschkin, Clark, de~Las~Casas, Guy, Jones, Bradbury, Johnson, Hechtman, Weidinger, Gabriel, Isaac, Lockhart, Osindero, Rimell, Dyer, Vinyals, Ayoub, Stanway, Bennett, Hassabis, Kavukcuoglu, and Irving]{rae2022scaling}
Jack~W. Rae, Sebastian Borgeaud, Trevor Cai, Katie Millican, Jordan Hoffmann, Francis Song, John Aslanides, Sarah Henderson, Roman Ring, Susannah Young, Eliza Rutherford, Tom Hennigan, Jacob Menick, Albin Cassirer, Richard Powell, George van~den Driessche, Lisa~Anne Hendricks, Maribeth Rauh, Po-Sen Huang, Amelia Glaese, Johannes Welbl, Sumanth Dathathri, Saffron Huang, Jonathan Uesato, John Mellor, Irina Higgins, Antonia Creswell, Nat McAleese, Amy Wu, Erich Elsen, Siddhant Jayakumar, Elena Buchatskaya, David Budden, Esme Sutherland, Karen Simonyan, Michela Paganini, Laurent Sifre, Lena Martens, Xiang~Lorraine Li, Adhiguna Kuncoro, Aida Nematzadeh, Elena Gribovskaya, Domenic Donato, Angeliki Lazaridou, Arthur Mensch, Jean-Baptiste Lespiau, Maria Tsimpoukelli, Nikolai Grigorev, Doug Fritz, Thibault Sottiaux, Mantas Pajarskas, Toby Pohlen, Zhitao Gong, Daniel Toyama, Cyprien de~Masson~d'Autume, Yujia Li, Tayfun Terzi, Vladimir Mikulik, Igor Babuschkin, Aidan Clark, Diego de~Las~Casas, Aurelia Guy, Chris Jones,
  James Bradbury, Matthew Johnson, Blake Hechtman, Laura Weidinger, Iason Gabriel, William Isaac, Ed~Lockhart, Simon Osindero, Laura Rimell, Chris Dyer, Oriol Vinyals, Kareem Ayoub, Jeff Stanway, Lorrayne Bennett, Demis Hassabis, Koray Kavukcuoglu, and Geoffrey Irving.
\newblock Scaling language models: Methods, analysis \& insights from training gopher, 2022.

\bibitem[Ruder(2022)]{ruder2023progress}
Sebastian Ruder.
\newblock Nlp-progress, 02 2022.
\newblock URL \url{https://nlpprogress.com/}.

\bibitem[Ruder \& Plank(2018)Ruder and Plank]{ruder-plank-2018-strong}
Sebastian Ruder and Barbara Plank.
\newblock Strong baselines for neural semi-supervised learning under domain shift.
\newblock In \emph{Proceedings of the 56th Annual Meeting of the Association for Computational Linguistics (Volume 1: Long Papers)}, pp.\  1044--1054, Melbourne, Australia, July 2018. Association for Computational Linguistics.
\newblock \doi{10.18653/v1/P18-1096}.
\newblock URL \url{https://aclanthology.org/P18-1096}.

\bibitem[Thoppilan et~al.(2022)Thoppilan, Freitas, Hall, Shazeer, Kulshreshtha, Cheng, Jin, Bos, Baker, Du, Li, Lee, Zheng, Ghafouri, Menegali, Huang, Krikun, Lepikhin, Qin, Chen, Xu, Chen, Roberts, Bosma, Zhao, Zhou, Chang, Krivokon, Rusch, Pickett, Srinivasan, Man, Meier-Hellstern, Morris, Doshi, Santos, Duke, Soraker, Zevenbergen, Prabhakaran, Diaz, Hutchinson, Olson, Molina, Hoffman-John, Lee, Aroyo, Rajakumar, Butryna, Lamm, Kuzmina, Fenton, Cohen, Bernstein, Kurzweil, Aguera-Arcas, Cui, Croak, Chi, and Le]{thoppilan2022lamda}
Romal Thoppilan, Daniel~De Freitas, Jamie Hall, Noam Shazeer, Apoorv Kulshreshtha, Heng-Tze Cheng, Alicia Jin, Taylor Bos, Leslie Baker, Yu~Du, YaGuang Li, Hongrae Lee, Huaixiu~Steven Zheng, Amin Ghafouri, Marcelo Menegali, Yanping Huang, Maxim Krikun, Dmitry Lepikhin, James Qin, Dehao Chen, Yuanzhong Xu, Zhifeng Chen, Adam Roberts, Maarten Bosma, Vincent Zhao, Yanqi Zhou, Chung-Ching Chang, Igor Krivokon, Will Rusch, Marc Pickett, Pranesh Srinivasan, Laichee Man, Kathleen Meier-Hellstern, Meredith~Ringel Morris, Tulsee Doshi, Renelito~Delos Santos, Toju Duke, Johnny Soraker, Ben Zevenbergen, Vinodkumar Prabhakaran, Mark Diaz, Ben Hutchinson, Kristen Olson, Alejandra Molina, Erin Hoffman-John, Josh Lee, Lora Aroyo, Ravi Rajakumar, Alena Butryna, Matthew Lamm, Viktoriya Kuzmina, Joe Fenton, Aaron Cohen, Rachel Bernstein, Ray Kurzweil, Blaise Aguera-Arcas, Claire Cui, Marian Croak, Ed~Chi, and Quoc Le.
\newblock Lamda: Language models for dialog applications, 2022.

\bibitem[Tjong Kim~Sang \& De~Meulder(2003)Tjong Kim~Sang and De~Meulder]{tjong-kim-sang-de-meulder-2003-introduction}
Erik~F. Tjong Kim~Sang and Fien De~Meulder.
\newblock Introduction to the {C}o{NLL}-2003 shared task: Language-independent named entity recognition.
\newblock In \emph{Proceedings of the Seventh Conference on Natural Language Learning at {HLT}-{NAACL} 2003}, pp.\  142--147, 2003.
\newblock URL \url{https://aclanthology.org/W03-0419}.

\bibitem[Vaswani et~al.(2017)Vaswani, Shazeer, Parmar, Uszkoreit, Jones, Gomez, Kaiser, and Polosukhin]{vaswani2017attention}
Ashish Vaswani, Noam Shazeer, Niki Parmar, Jakob Uszkoreit, Llion Jones, Aidan~N Gomez, {\L}ukasz Kaiser, and Illia Polosukhin.
\newblock Attention is all you need.
\newblock \emph{Advances in Neural Information Processing Systems}, 30:\penalty0 5998--6008, 2017.

\bibitem[Warstadt et~al.(2023)Warstadt, Choshen, Mueller, Williams, Wilcox, and Zhuang]{warstadt2023papers}
Alex Warstadt, Leshem Choshen, Aaron Mueller, Adina Williams, Ethan Wilcox, and Chengxu Zhuang.
\newblock Call for papers -- the babylm challenge: Sample-efficient pretraining on a developmentally plausible corpus.
\newblock \emph{Computing Research Repository}, arXiv:2301.11796, 2023.

\bibitem[Xu et~al.(2015)Xu, Wang, Chen, and Li]{xu2015empirical}
Bing Xu, Naiyan Wang, Tianqi Chen, and Mu~Li.
\newblock Empirical evaluation of rectified activations in convolutional network, 2015.

\bibitem[Zeldes(2017)]{Zeldes2017}
Amir Zeldes.
\newblock The {GUM} corpus: Creating multilayer resources in the classroom.
\newblock \emph{Language Resources and Evaluation}, 51\penalty0 (3):\penalty0 581--612, 2017.
\newblock \doi{http://dx.doi.org/10.1007/s10579-016-9343-x}.

\end{thebibliography}
\bibliographystyle{iclr2024_conference}
\clearpage

\appendix
\section{Appendix}
\label{sec:appendix}
In appendix, we documented all the probing experiments results of all the bit-cipher models we trained both on POS tagging and NER with numbers in the tables are shown as accuracy with F1-scores shown in parentheses.

\begin{table*}[!h]
\caption{Table for 60 \textbf{Sum} models of bit-cipher on POS tagging probing experiments}\label{tab:sum-pos}
\centering
\label{tab:sum-pos}

\small 
\renewcommand{\arraystretch}{1.5} 

\begin{tabularx}{\textwidth}{lXXXXXXXXXXXX}
\toprule
\textbf{bits} & \multicolumn{3}{c}{b = 25} & \multicolumn{3}{c}{b = 50} & \multicolumn{3}{c}{b = 100} & \multicolumn{3}{c}{b = 200} \\
\cmidrule(lr){2-4} \cmidrule(lr){5-7} \cmidrule(lr){8-10} \cmidrule(lr){11-13}
\textbf{DataSize} & \textbf{r = 1} & \textbf{r = 2} & \textbf{r = 4} & \textbf{r = 1} & \textbf{r = 2} & \textbf{r = 4} & \textbf{r = 1} & \textbf{r = 2} & \textbf{r = 4} & \textbf{r = 1} & \textbf{r = 2} & \textbf{r = 4}  \\
\midrule
0.5B & 82.91 (83.08) & 82.70 (82.73)& 81.23 (80.85)&  84.35 (84.62)& 83.67 (84.02)& 83.17 (83.13)& 84.42 (84.52)& 84.67 (84.97)& 83.67 (83.91)& 84.89 (85.18)& 84.64 (84.87)& 84.21 (84.50)\\
1.0B & 82.87 (83.04)& 82.36 (82.23)& \textbf{82.25 (82.21)}& 84.52 (84.72)& 83.88 (83.95)& 83.52 (83.67)& \textbf{85.89 (86.19)}& 84.97 (84.92)& 84.25 (84.32)& \textbf{85.43 (85.65)}& 85.10 (85.40)&  84.81 (85.04)\\
2.0B & 83.50 (83.78)& 82.91 (82.89)& 81.75 (81.40)& 84.55 (84.71)&  84.01 (84.18)& 83.25  (83.37)& 85.43 (85.70)& 84.92 (85.04)& 84.30 (84.47)& 85.42 (85.71)& 85.63 (85.92)& 85.45 (85.71)\\
4.0B & 83.83 (83.87)& \textbf{83.07 (83.08)}&  82.00 (81.87)& \textbf{84.85 (85.01)}& 84.32 (84.50)& 83.53 (83.70)& 85.65 (85.90)& \textbf{85.64 (85.90)}& 84.61 (85.00)& 85.18 (85.49)& 84.83 (84.87)&  85.12 (85.40)\\
8.0B & \textbf{83.93 (84.08)}& 82.85 (82.74) & 82.17 (81.96) & 84.44 (84.68)& \textbf{84.34 (84.46)} & \textbf{83.83 (84.08)}& 84.77 (84.99) & 85.28 (85.52)& \textbf{84.85 (85.14)}& 85.36 (85.20)& \textbf{86.20 (86.47)}& \textbf{85.67 (86.04)}\\
\bottomrule
\end{tabularx}

\end{table*}

\begin{table*}[!h]
\caption{Table for 60 \textbf{Sum} models of bit-cipher on NER tagging probing experiments}
\centering
\label{tab:sum-ner}
\small 
\renewcommand{\arraystretch}{1.5} 
\begin{tabularx}{\textwidth}{lXXXXXXXXXXXX}
\toprule
\textbf{bits} & \multicolumn{3}{c}{b = 25} & \multicolumn{3}{c}{b = 50} & \multicolumn{3}{c}{b = 100} & \multicolumn{3}{c}{b = 200} \\
\cmidrule(lr){2-4} \cmidrule(lr){5-7} \cmidrule(lr){8-10} \cmidrule(lr){11-13}
\textbf{DataSize} & \textbf{r = 1} & \textbf{r = 2} & \textbf{r = 4} & \textbf{r = 1} & \textbf{r = 2} & \textbf{r = 4} & \textbf{r = 1} & \textbf{r = 2} & \textbf{r = 4} & \textbf{r = 1} & \textbf{r = 2} & \textbf{r = 4}  \\
\midrule
0.5B & \textbf{91.50 (91.17)} & 89.08 (89.24)& 89.12 (89.12)& 89.60 (89.95)& 89.53 (89.80)& 89.12 (89.12)& 89.59 (90.02)& 89.78 (90.38)& 89.94 (90.31)& 89.47 (90.03)& 89.74 (90.33)& \textbf{92.35 (92.35)}\\
1.0B & 89.44 (89.75)& 89.26 (89.31)& \textbf{89.46 (89.69)}& \textbf{92.23 (91.94)}& 89.78 (90.11)& 89.84 (90.13)& 89.97 (90.41)& 90.22 (90.60)& 90.20 (90.61)& 89.95 (90.44)& 90.03 (90.56)& 90.30 (90.92)\\
2.0B & 89.48 (89.74)& 89.58 (89.64)& 89.13 (88.75)& 90.04 (90.47)& 89.51 (89.55)& 89.93 (90.23)& 90.23 (90.81)& 90.38 (90.98)& 90.22 (90.70)& 90.20 (90.84)& 90.45 (90.91)& 90.50 (91.05)\\
4.0B & 89.74 (90.06)& \textbf{89.74 (90.01)}& 89.41 (89.05)& 90.25 (90.69)& \textbf{90.40 (90.82)}& 89.77 (90.10)& 90.31 (90.81)& 90.38 (90.96)& 90.60 (90.99)& 90.31 (91.01)& 0.9042 (91.06)& 90.64 (91.25)\\
8.0B & 89.97 (90.20)& 89.64 (89.56)& 89.32 (89.13)& 90.50 (91.03)& 90.31 (90.67)& \textbf{90.61 (91.07)}& \textbf{90.65 (91.24)}& \textbf{90.71 (91.30)}& \textbf{90.82 (91.35)}& \textbf{90.63 (91.27)}& \textbf{90.72 (91.25)}& 90.67 (91.32)\\
\bottomrule
\end{tabularx}
\end{table*}

\begin{table*}[!h]
\caption{Table for 20 \textbf{Cat} models of bit-cipher on NER tagging probing experiments}
\centering
\begin{tabular}{lcccc}
\hline
\textbf{bits} & \textbf{25b (200d)} & \textbf{50b (400d)} & \textbf{100b (800d)} & \textbf{200b (1600d)}\\
\textbf{data-size} \\
\hline
0.5B & 89.90 (90.48)& 89.93 (90.45)& 89.90 (90.48)& 89.83 (90.34)\\ 
1.0B & 90.24 (90.81)& 90.49 (90.93)& 90.31 (90.92)& 90.18 (90.61)\\ 
2.0B & 90.19 (90.49)& 90.42 (91.00)& 90.44 (91.02)& 90.28 (90.85)\\  
4.0B & 90.70 (91.22)& 90.74 (91.14)& 90.60 (91.22)& 90.49 (90.99)\\ 
8.0B & \textbf{90.96 (91.51)} & \textbf{90.91 (91.62)}& \textbf{90.80 (91.50)} & \textbf{90.81 (91.25)}\\\hline
\end{tabular}

\label{tab:cat-ner}
\end{table*}

\begin{table*}[!h]
\caption{Table for 20 \textbf{Cat} models of bit-cipher on POS tagging probing experiments}
\centering
\begin{tabular}{lcccc}
\hline
\textbf{bits} & \textbf{25b (200d)} & \textbf{50b (400d)} & \textbf{100b (800d)} & \textbf{200b (1600d)}\\
\textbf{data-size} \\
\hline
0.5B & 85.53 (85.88) & 86.00 (86.29) & 85.60 (85.81) & 85.40 (85.75)\\ 
1.0B & 85.81 (86.35)& 85.71 (85.89)& 86.13 (86.44)& 85.17(85.57)\\ 
2.0B & 85.93 (86.06)& 85.78 (86.24)& 85.97 (86.17) & 85.42 (85.88)\\  
4.0B & 85.48 (85.95)& 85.56 (85.74)& 85.92 (86.15)& 85.53 (86.04)\\ 
8.0B & \textbf{86.05 (86.32)}& \textbf{86.19 (86.63)}& \textbf{86.16 (86.48)}& \textbf{85.93 (86.20)}\\\hline
\end{tabular}

\label{tab:cat-pos}
\end{table*}

\begin{table*}[!h]
\caption{Table for 20 \textbf{Cip} models of cipher on its own on POS probing experiments}
\centering
\begin{tabular}{lcccc}
\hline
\textbf{bits} & \textbf{25b} & \textbf{50b} & \textbf{100b} & \textbf{200b}\\
\textbf{data-size} \\
\hline
0.5B & 73.76 (71.43) & 74.77 (73.08)& \textbf{75.31 (73.43)} & 75.21(73.56)\\ 
1.0B & 73.65 (71.26) & 74.27 (72.44)& 74.64 (73.21) & \textbf{75.86 (73.92)}\\ 
2.0B & \textbf{73.89 (71.50)} & 74.93 (72.94)& 75.49 (73.82) & 75.69 (73.90)\\  
4.0B & 72.21 (69.63) & 74.80 (73.06)& 74.93 (73.21)& 75.26 (73.68)\\ 
8.0B & 72.72 (70.44) & \textbf{75.02 (73.41)} & 75.22 (73.46)& 75.23 (73.58)\\\hline
\end{tabular}

\label{tab:cip-pos}
\end{table*}

\begin{table*}[!h]
\caption{Table for 20 \textbf{Cip} models of cipher on its own on NER probing experiments}
\centering
\begin{tabular}{lcccc}
\hline
\textbf{bits} & \textbf{25b} & \textbf{50b} & \textbf{100b} & \textbf{200b}\\
\textbf{data-size} \\
\hline
0.5B & 85.02 (83.55) & 85.64 (83.97)& 85.80 (83.92) & 85.83 (83.90)\\ 
1.0B & 84.20 (82.72)& 85.58 (83.64)& 85.68 (83.61) & 85.75 (83.84)\\ 
2.0B & 82.76 (82.20)& \textbf{85.75 (83.97)}& 85.75 (83.82) & 86.04 (84.14)\\  
4.0B & \textbf{85.22 (83.85)}& 85.66 (83.99)& 84.83 (83.50)& 85.98 (84.14)\\ 
8.0B & 85.17 (83.83)& 85.54 (83.93)& \textbf{85.90 (84.03)}& \textbf{86.19 (84.17)}\\\hline
\end{tabular}
\label{tab:cip-ner}
\end{table*}

\end{document}